\documentclass[a4paper, 10pt, conference]{ieeeconf}      % Use this line for a4
\IEEEoverridecommandlockouts                              % This command is only
\usepackage[english]{babel}
\usepackage[utf8]{inputenc}
\usepackage{algorithm}
\usepackage{algpseudocode}
\usepackage{graphicx}
\usepackage[draft]{minted}

\title{\LARGE \bf
Visual Servoing of Unmanned Surface Vehicle from Small Tethered Unmanned Aerial Vehicle
}

\author{Aritra Biswas$^{1}$, Haresh Karnan$^{1}$, Pranav Vaidik Dhulipala$^{2}$, Jan Dufek$^{3}$, and Robin Murphy$^{3}$% <-this % stops a space
\thanks{$^{1}$A. Biswas and H. Karnan are with the Department of Aerospace Engineering, Texas A\&M University,
        College Station, TX 77840, USA
        {\tt\small aritra@tamu.edu}{\tt\small , haresh.miriyala@tamu.edu} }
\thanks{$^{2}$P. V. Dhulipala is with the Department of Electrical Engineering, Texas A\&M University,
        College Station, TX 77840, USA
        {\tt\small pranav.d1993@tamu.edu}}%
\thanks{$^{3}$J. Dufek and Dr.Robin Murphy are with the Department of Computer Science and Engineering, Texas A\&M University,
        College Station, TX 77840, USA
        {\tt\small robin.r.murphy@tamu.edu}{\tt\small, dufek@tamu.edu}}           
}

\begin{document}

\maketitle
\thispagestyle{empty}
\pagestyle{empty}

%%%%%%%%%%%%%%%%%%%%%%%%%%%%%%%%%%%%%%%%%%%%%%%%%%%%%%%%%%%%%%%%%%%%%%%%%%%%%%%%
\begin{abstract}
This paper presents an algorithm and the implementation of a motor schema to aid the visual localization subsystem of the ongoing EMILY project at Texas A\&M University. The EMILY project aims to team an Unmanned Surface Vehicle (USV) with an Unmanned Aerial Vehicle (UAV) to augment the search and rescue of marine casualties during an emergency response phase. The USV is designed to serve as a flotation device once it reaches the victims. A live video feed from the UAV is provided to the casuality responders, giving them a visual estimate of the USV's orientation and position to help with its navigation. One of the challenges involved with casualty response using a USV-UAV team is to simultaneously control the USV and track it. In this paper, we present an implemented solution to automate the UAV camera movements to keep the USV in view at all times. The motor schema proposed, uses the USV's coordinates from the visual localization subsystem to control the UAV's camera movements and track the USV with minimal camera movements, such that the USV is always in the camera's field of view. The motor schema is validated in two proof-of-concept laboratory trials and the results are discussed in the later sections.

\end{abstract}

%%%%%%%%%%%%%%%%%%%%%%%%%%%%%%%%%%%%%%%%%%%%%%%%%%%%%%%%%%%%%%%%%%%%%%%%%%%%%%%%
\section{Introduction}

Refugees from war torn countries like Syria and Libya are often found making their escape though Mediterranean sea in hopes of finding asylum in Europe. Many a times they find themselves in a refugee boat that is overcrowded with fellow refugees, in poor conditions, and sometimes even with a broken rudder \cite{c5}. As a result, there have been several cases of drowned victims either due to the boats meeting tragic fates at the sea, or due to the victims falling off from the boats. 

Rescue boats that assist the refugees may find it challenging to reach all the victims due to their limited number and difficulty in accessing shallow waters or regions near the cliffs. Also, the individual responders cannot go too far from the shore due to lack of adequate technology. 

The use of USVs for disaster management robotics has become an interesting and a practical application during the recent years. There have been multiple instances since 2004 \cite{c2} where USVs were used for disaster recovery operations. USVs were deployed for recovery operations in situations like hurricanes \cite{c3} \cite{c4} \cite{c11} and tsunamis \cite{c12} in the past, but the primary focus of their application has not been in the response phase of the disaster management.

In the cases of search and rescue operations of drowning victims, it is critical to have real-time responsiveness. USVs can be deployed in these scenarios, as they can reach the drowning victims faster than individual responders at sea. They are also much more maneuverable than the rescue boats due to their smaller size and their reachability in shallow waters.

The Emergency Integrated Lifesaving Lanyard (EMILY) project by Hydronalix \cite{c8}, proposes to use a USV to assist the rescuers to reach the drowning victims and rescue them. The USV is able to act as a flotation device that can support up to 6 people and was tested by Murphy et al., in teleoperation mode at Lesbos, Greece in 2016. The USV is designed not to sink when the victims are holding onto it, buying the rescuers enough time to reach them. 

The ongoing autonomous EMILY project aims to team an autonomous USV with a UAV. The idea is to have the UAV provide the operator with live video feed of EMILY's location and orientation and also to help locate the victims. Thus the responders would be able to send EMILY to the victims on the water body while they could help the rescued with more urgent issues in person.

The visual localization subsystem \cite{c1} was proposed by Dufek et al., for the project in 2016, to help find EMILY's location and orientation from the UAV's video feed. This subsystem aids the operator in finding the USV's orientation and location easily by letting the user manually select the USV and the target from the video feed, during which the pose is automatically estimated. A PID controller then uses the pose data of the USV and reduces the distance between the USV and target to zero, by steering the USV to reach the target. One disadvantage that still persists with the subsystem is that the operator will have to manually control the camera orientations in the UAV to locate EMILY whenever it leaves the field of view. This forces the user to operate the UAV and USV simultaneously, limiting the efficiency of the user in controlling the USV.

This paper proposes to use a motor schema behavior in conjunction with the visual localization subsystem. The motor schema autonomously controls the yaw and pitch movements of the UAV camera so that the USV always stays in the camera's field of view thereby eliminating the user's work of controlling the camera orientation.

\section{Related Work}

Previous work by Dufek et al. shows the evaluation of algorithms for the visual localization subsystem and establishes the histogram based approach to be the most effective in determining the location and orientation of the USV from the UAV's video feed \cite{c1}. This paper focuses on the extension of Dufek's work by adding an autonomous system to control the camera's pan/tilt to track the USV. In this section, we evaluate multiple techniques to use pan-tilt movements for visual servoing.

A camera-control approach by UAV was proposed in \cite{c15}, where the UAV camera movements are based on a biomimetic eye based control, being able to simulate the human eye movements resulting in stable and smooth camera movements. The algorithm is computationally heavy, which is not suitable for our purposes.

Kang et al., \cite{c14} proposes to use a PTZ camera for tracking objects and humans in controlled lighting conditions. However, camera-control algorithm tries to center the object in the frame which would make camera happen little too often and consumes more power, which is one of the constraints of the project.

YongChol et al., in \cite{c7} shows the usage of a camera and PIR sensor fusion to track a moving object (human). However, this method has a range limitation posed by the Passive Infrared sensor. For the UAV to track the USV, an algorithm that can track the USV irrespective of the distance of the UAV from the USV, as long as it is in sight of the camera is needed. This is because, the UAV is not stationary as it keeps moving due to the effects of wind and coastal currents in the sea. Also, the PIR sensor consumes additional power and is not reliable in sunlight.

Use of encoders in the camera for feedback position control as in \cite{c8} is not advisable because encoders consume additional power from the UAV's battery. Since battery time is limited for a UAV, it is best to use limited number of devices.

As opposed to the aforementioned approaches, this paper focuses on a rather simple algorithm, with an asymptotic complexity of O(1) to facilitate the real-time tracking control. Also, the algorithm focuses on moving the USV's position into the Elliptical Region of Interest (ROI) and only moves the camera when the USV's location in the video feed moves out of the ellipse but stays put otherwise. An advantage of having a faster algorithm is that it can be run real-time on an onboard computer on the UAV. Studies and laboratory trials as discussed in the further sections show that the number of camera movements and hence the power consumption are minimum with the use of this algorithm.

\section{Approach}

The multi-robot system discussed in this research consists of an USV (EMILY) and a Fotokite UAV. The UAV provides visual feedback either to the human responders for manually navigating EMILY or to the autonomous visual navigation subsystem, as developed by Xuesu et al., From the video stream, EMILY's position is identified using a histogram based Camshift and Back-projection algorithm as discussed by Dufek et al., in \cite{c1}.\par
\begin{figure}[h!]
  \centering
  \includegraphics[width=0.5\textwidth]{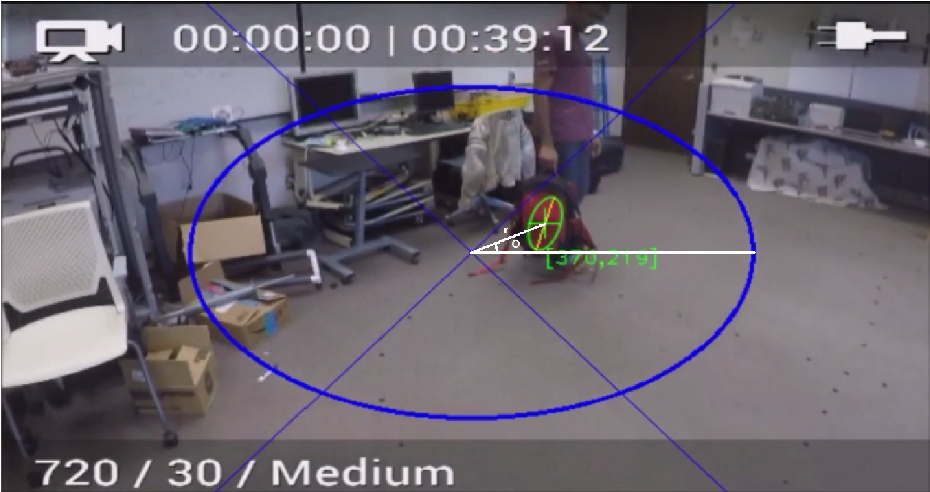}
  \caption{Elliptical ROI in blue, with $\theta,r$ denoted in white.}
  \label{fig:screencapture}
\end{figure}

 An elliptical ROI is introduced onto the image frame as shown in Fig~\ref{fig:screencapture} inside of which EMILY will be placed by controlling either the pitch or yaw of the camera. The major and minor axis of the ellipse are chosen to be a preset percentage of the length and height of the image frame. The boundary of the ROI is chosen to be an ellipse, owing to the simplicity that this shape provides in computing relative position of a point to its center. The relative position of EMILY, $P$, is calculated using the simple equation : 
\begin{equation} \label{pval}
P = \frac{x^2}{a^2}+\frac{y^2}{b^2}
\end{equation} 
where $x$ and $y$ are the x,y pixel values of EMILY in the camera's frame and $a,b$ are the major and the minor axis of the elliptical ROI in the image axis where the origin is at the center of the image. If P is greater than 1, EMILY is outside the ROI. Otherwise, it is inside the ROI or on the surface of the ellipse. When P is greater than 1, the controller tries to move either the pitch of the Camera or yaw of the Fotokite. If P is equal to or lesser than 1, the camera is not moved. 

\begin{figure}[h!]
  \centering
  \includegraphics[width=0.5\textwidth]{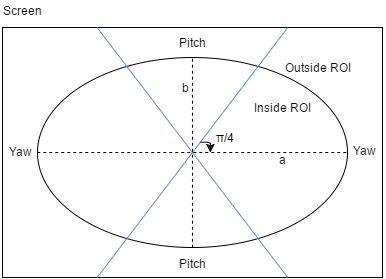}
  \caption{Sector representation of the ellipse}
  \label{fig:sector}
\end{figure}

First, the cartesian coordinates x,y are converted into polar coordinates r (distance) and $\theta$ (tangent angle) as shown in Fig~\ref{fig:screencapture} and the relative distance P is computed. When the relative distance becomes greater than 1, the control code acts to move the pitch or yaw to bring EMILY into the ROI. The ellipse is further divided into 4 sectors $left, right, top$, and $bottom$ as shown in Fig~\ref{fig:sector}. The two sector division lines (represented blue in Fig~\ref{fig:sector}) make an angle of $\pi$/4 with the major axis of the ellipse as shown in Fig~\ref{fig:sector} .The control algorithm works such that a pitch magnitude command is sent when EMILY moves outside the ROI either in the top or bottom sectors. Similarly, a yaw magnitude command is sent only when EMILY is either in the left or the right sectors outside the ROI. 
\begin{figure}[h!]
  \centering
  \includegraphics[width=0.5\textwidth]{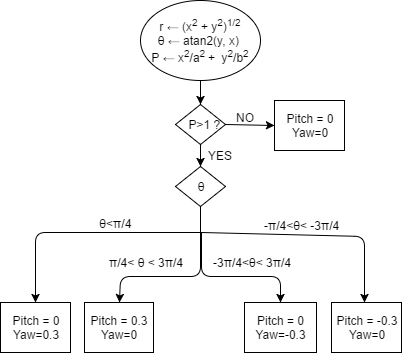}
  \caption{Flowchart of the control algorithm}
  \label{fig:algoflowchart}
\end{figure}

A pseudo code representation of the algorithm discussed is shown in Algorithm 1 and a flowchart representation of the same is shown in Fig~\ref{fig:algoflowchart}.

\begin{algorithm}
\caption{Control algorithm}
\begin{algorithmic}
\Procedure{Control Code}{$x,y,a,b$}\Comment{$x,y$ coordinates of EMILY; $a,b$ major and minor axis lengths of the ellipse}
\State $r\gets \sqrt[2]{x^2 + y^2} $ 
\State $\theta \gets atan2(y,x) $ 
\State $P\gets \frac{x^2}{a^2} + \frac{y^2}{b^2} $
\If{$P$ is greater than 1}
\If{$y$ is less than 0}
\State $r\gets -r $ 
\EndIf
\If{$\theta$ is less than $\pi$/4}
\State $Yaw\gets 0.3, Pitch \gets 0 $ 

\ElsIf {$\theta$ is between -3$\pi$/4 and 3$\pi$/4}
\State $Yaw\gets -0.3, Pitch \gets 0 $ 

\ElsIf {$\theta$ is between $\pi$/4 and 3$\pi$/4}
\State $Yaw\gets 0, Pitch \gets 0.3 $ 

\ElsIf {$\theta$ is between -$\pi$/4 and -3$\pi$/4}
\State $Yaw\gets 0, Pitch \gets -0.3 $

\EndIf

\Else
\State $Yaw\gets 0, Pitch \gets 0 $ 
\EndIf

\EndProcedure
\end{algorithmic}
\end{algorithm}

\begin{figure}[h!]
  \centering
  \includegraphics[width=0.5\textwidth]{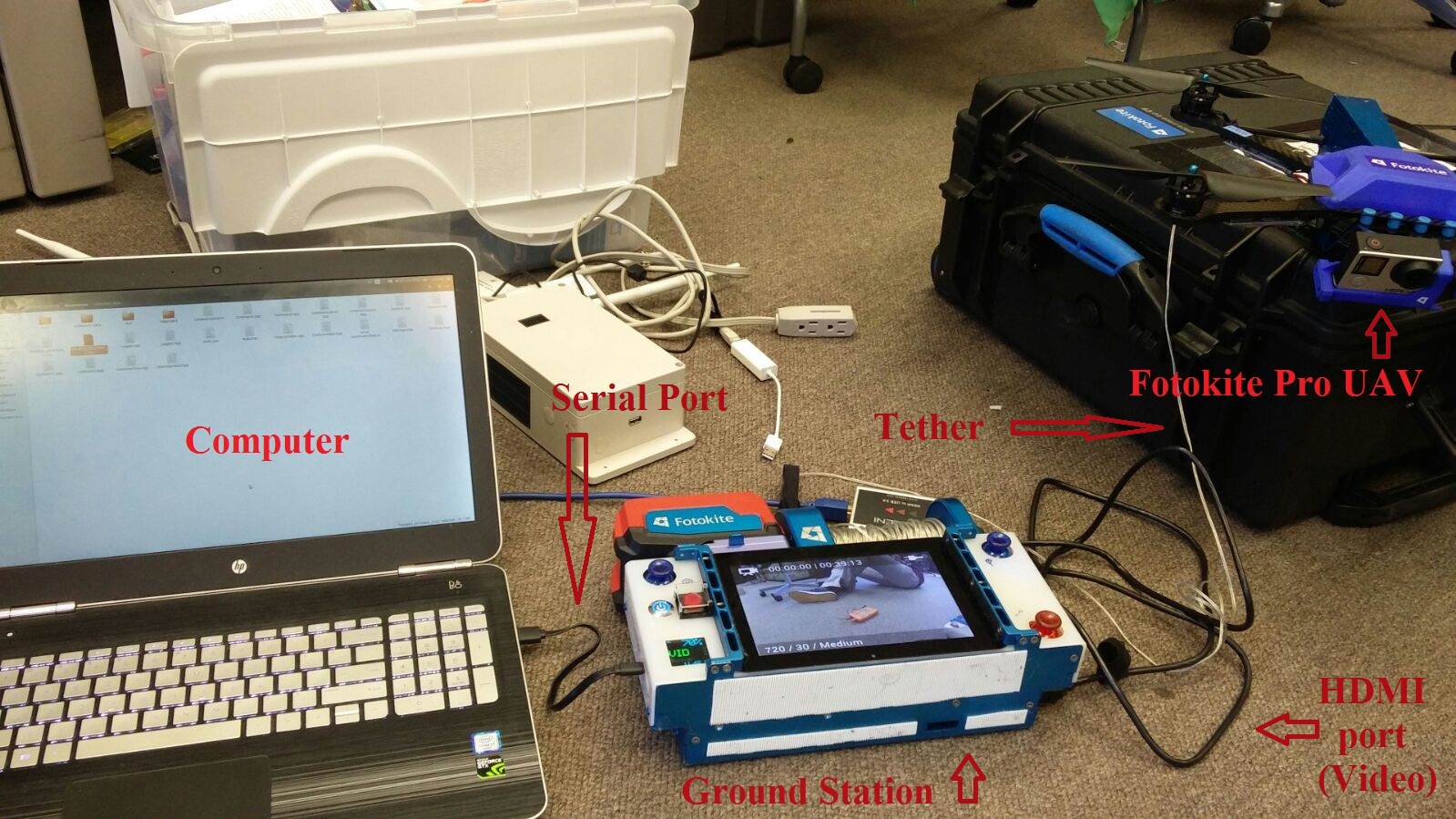}
  \caption{Experimental setup of Fotokite UAV}
  \label{fig:setupphoto}
\end{figure}

\begin{figure}[h!]
  \centering
  \includegraphics[width=0.5\textwidth]{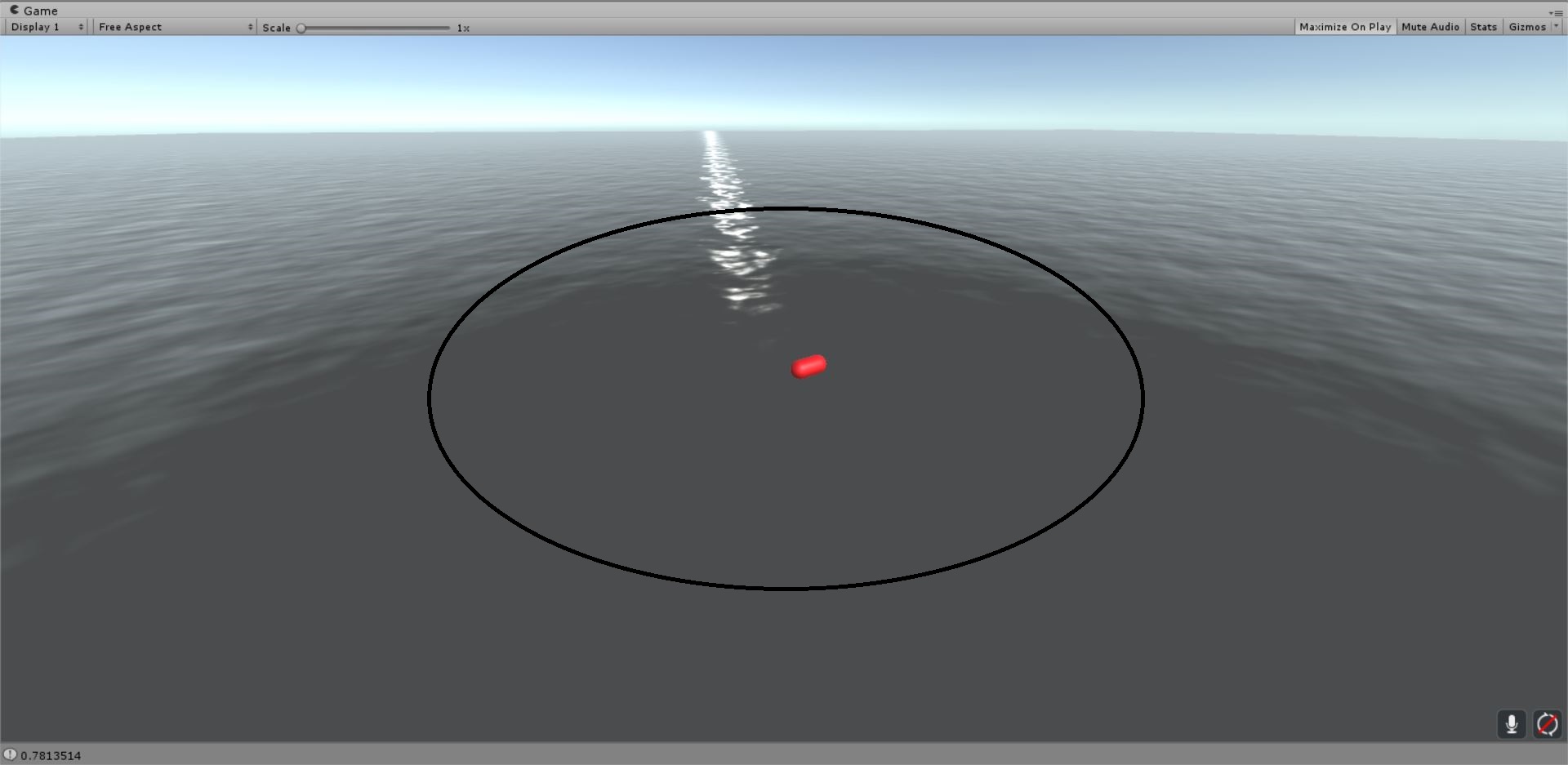}
  \caption{Simulation in UNITY 3D}
  \label{fig:unitysim}
\end{figure}
\begin{figure}[h!]
  \centering
  \includegraphics[width=0.5\textwidth]{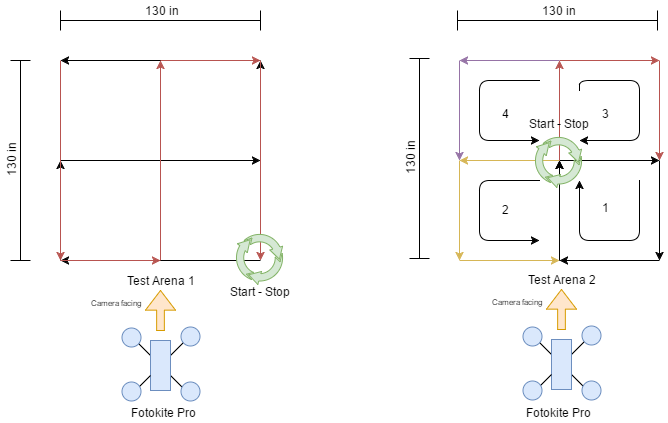}
  \caption{Two test arenas used for proof of concept trials}
  \label{fig:testarenas}
\end{figure}

\begin{figure}[h!]
  \centering
  \includegraphics[width=0.5\textwidth]{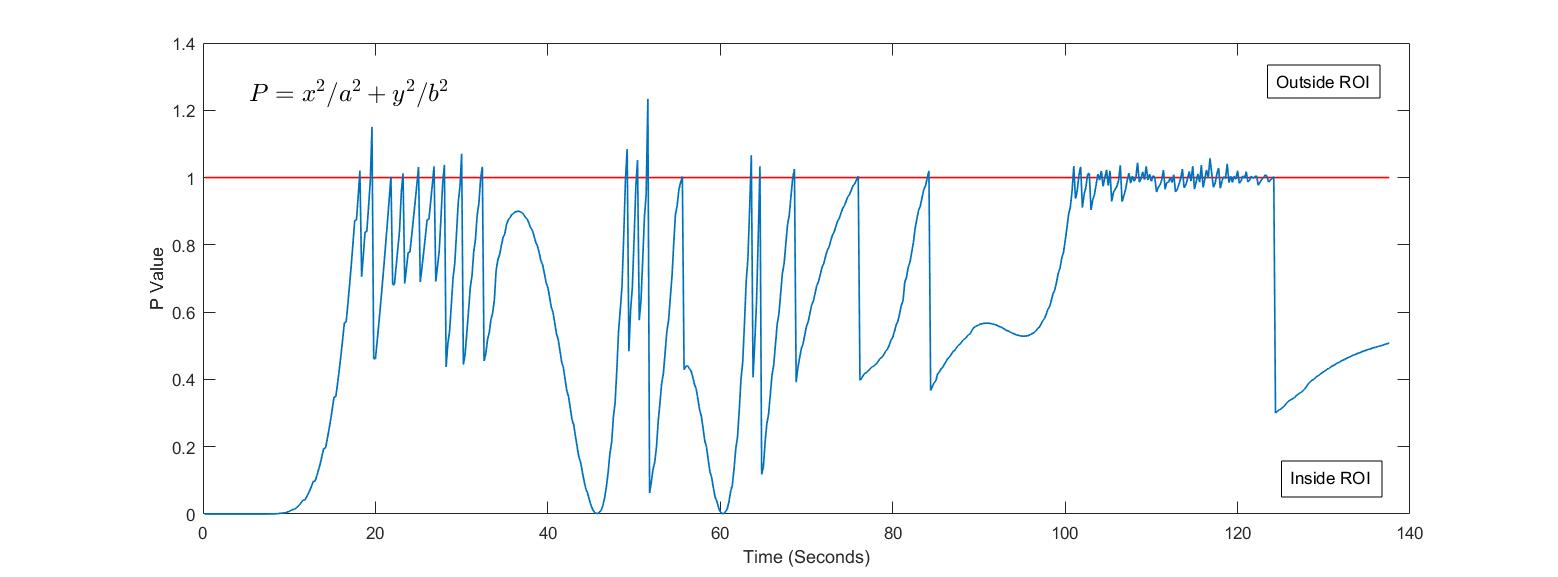}
  \caption{Results from UNITY 3D simulation}
  \label{fig:simgraph}
\end{figure}

\section{Implementations}
\subsection{Simulation of the tracking algorithm in UNITY 3D}
A simulation scene consisting of a bright red colored USV in water, with a camera placed on a UAV was developed in UNITY 3D. Fig~\ref{fig:unitysim} shows a frame from the UAV's video feed. The USV is modeled to be a holonomic vehicle, like EMILY, which can only move forward using thrust, and steer using its rudder. Directional arrows Up, Left, and Right are programmed to control the motion of the USV. The camera is able to perform pan/tilt motion relative to the UAV. The algorithm is tested for two test arena tracks, as shown in Fig~\ref{fig:testarenas}. These test cases are chosen because they maximize the instances when the USV goes out of the ROI. A simulation is run with 20 trials to validate the algorithm. Fig~\ref{fig:simgraph} shows the time history of the P value calculated as shown in equation \ref{pval}. The red line in Fig~\ref{fig:simgraph} represents the boundary of the ellipse. It can be seen that as the USV moves out of the elliptical boundary, the control algorithm moves the USV inside the ROI by controlling the pan or tilt of the camera.
\begin{figure}[h!]
  \centering
  \includegraphics[width=0.5\textwidth]{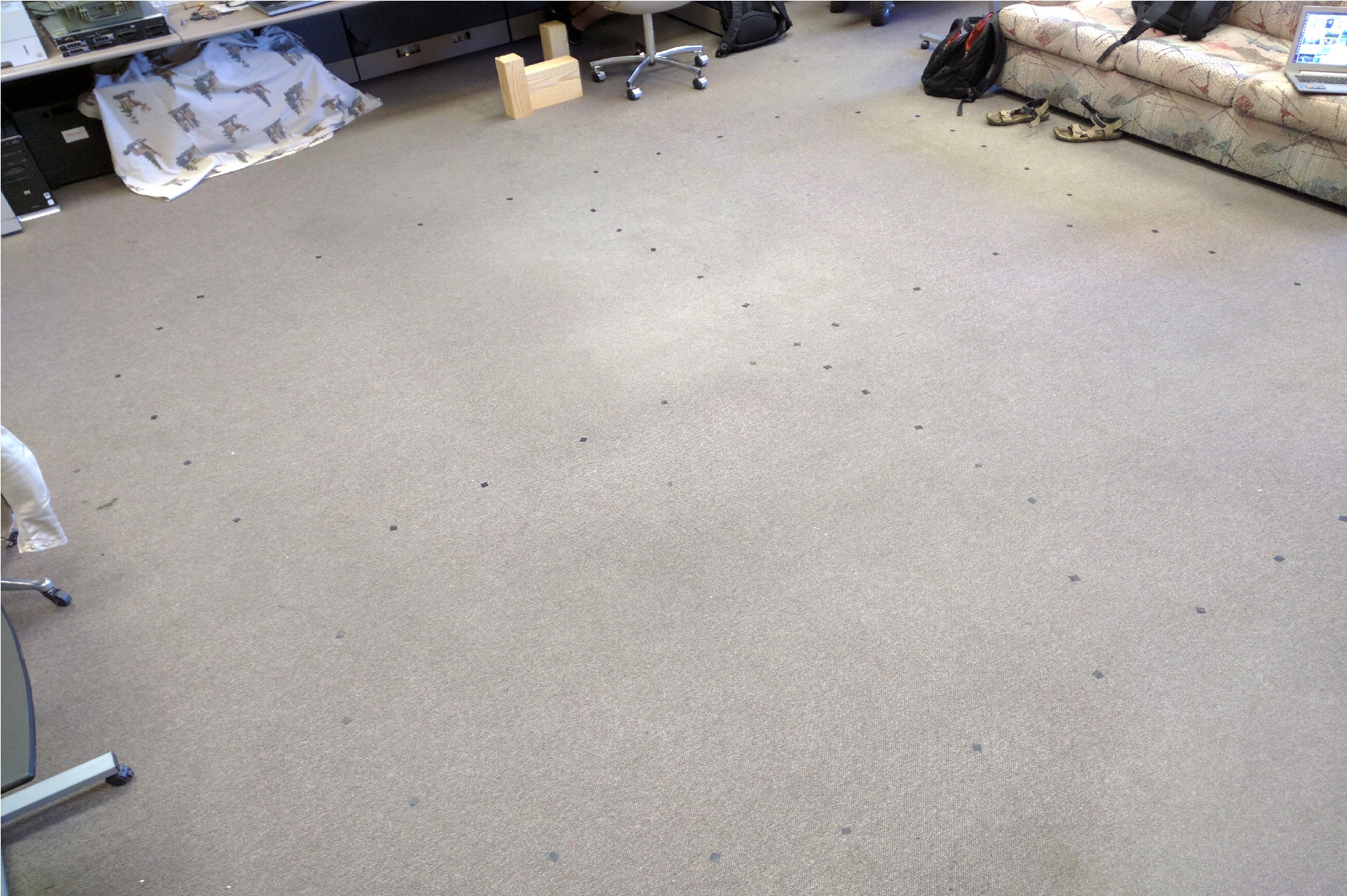}
  \caption{Arena 1 marked on the ground with black markers.}
  \label{fig:arenaphoto}
\end{figure}

\subsection{Laboratory test}
A Fotokite Pro is used as the UAV and a bright red object is chosen to substitute for EMILY. Since this algorithm uses only the position data of the object and not the orientation, any red object could be used for the test. Fotokite is chosen as the UAV because the UAV unit is tethered and receives uninterrupted power supply from the groundstation that can be powered from an AC supply, enabling extended flight times. Each unit (UAV and ground-station) has its own dedicated backup power supply from LiPo batteries. The Fotokite Pro has a gimbal equipped visual system installed on its frame. The camera used is a GoPro Hero 4 and it streams 720p HD video of 1920x720 frame size to the ground station and the computer.\par
To control the pitch or yaw of the UAV's camera, we use UART serial communication to talk to the groundstation-Fotokite system and send commands to it. Fotokite Pro accepts inputs for pitch or yaw in specific String formats with the angular velocity value in rad/sec appended to the end of the command. For example, the yaw can be controlled by sending "Yaw 0.2". The actuator used in the gimbal can move at a maximum of 0.3 rad/sec and so does the yaw control. \par
\begin{figure}[h!]
  \centering
  \includegraphics[width=0.5\textwidth]{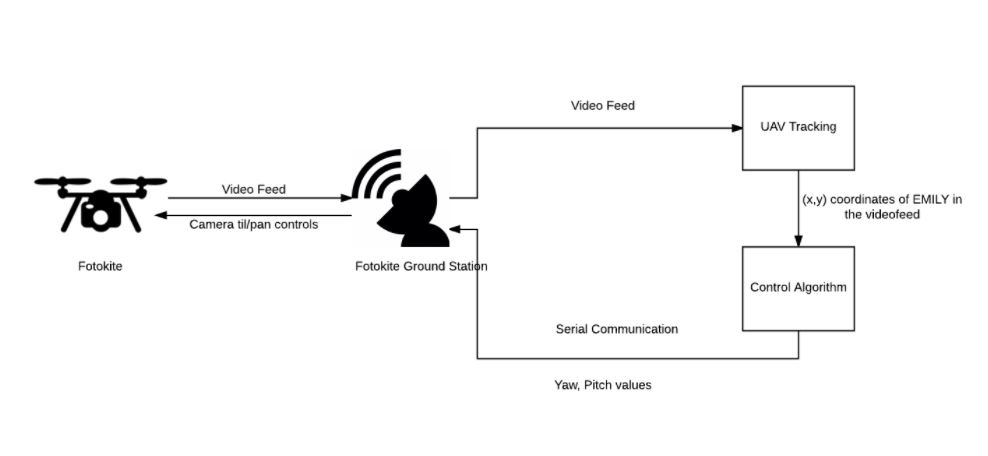}
  \caption{Flowchart of the Camera Control Implementation}
  \label{fig:cameraflowchart}
\end{figure}

The video feed from the camera is fed to the visual localization subsystem that generates the $x,y$ coordinates of the tracked object. Using this data, the relative position value $P$ is computed as in equation~\ref{pval}. The control acts only when $P$ is greater than 1 and generates yaw and pitch values. The generated yaw and pitch values are sent to the Fotokite ground station via a USART serial interface at a baudrate of 9600 bps. The experimental setup is shown in Fig~\ref{fig:setupphoto} and Fig~\ref{fig:cameraflowchart} shows the flowchart of the implementation.

\section{Trials}

Two test arenas are introduced and used in the testing phase to validate the algorithm. Fig~\ref{fig:testarenas} shows the two test areans - 1 and 2 and the respective paths. Fig~\ref{fig:arenaphoto} shows the setup of the arena in the laboratory space. The Fotokite is kept in flight mode at a constant altitude of 6 feet from the ground, with the control algorithm running real-time and a human operator moves the red object along the designated path in the arenas once every trial. The program records the location coordinates of the object, relative distance P from the center of the ROI, pitch and yaw control values over time.

The time taken for the camera to bring the object of interest back into the ROI is a measure of sensitivity of the algorithm and hardware combined. It is defined as follows:
\begin{equation} \label{pval2}
s = \frac{h}{b}
\end{equation}
where $s$ is the sensitivity, $h$ and $b$ are the height and breadth respectively of the peaks in the graph shown in Fig~\ref{fig:area1data}

To obtain a normalized metric independent of the arena chosen, the sensitivity metric shown above is divided by a value $n$, where $n$ refers to the number of times the object goes out of the ROI. In these two test cases, the values of n are 18 and 13 respectively in arena 1 and 2.

\subsection{Arena 1}

\begin{figure}[h!]
  \centering
  \includegraphics[width=0.5\textwidth]{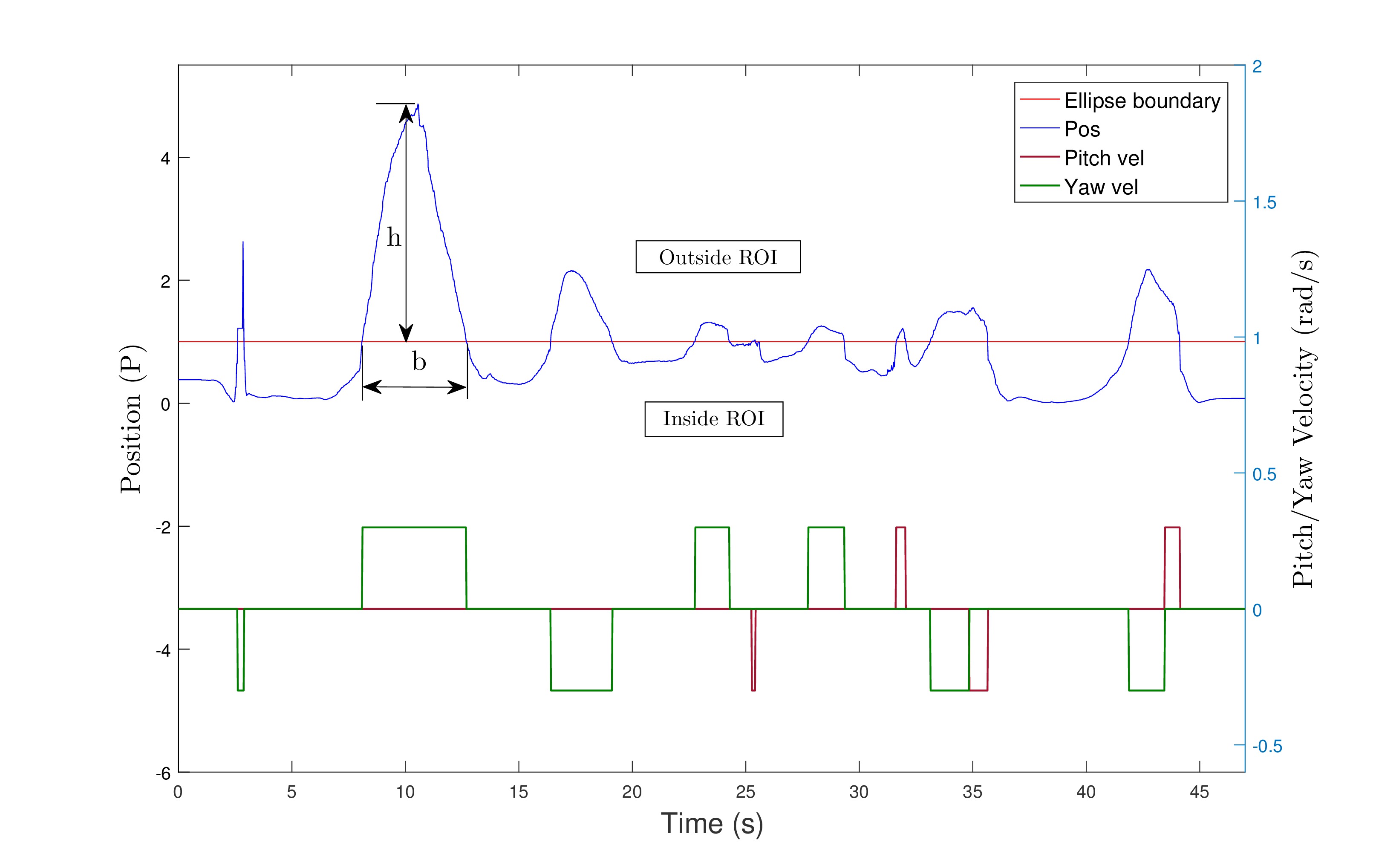}
  \caption{Time history of recorded data - Arena 1}
  \label{fig:area1data}
\end{figure}

13 trials were performed on this test arena and all were successful in tracking the red object. The relative distance from the center of ROI in the video feed from the boundary of the ellipse is recorded and is termed here as P. Plotting the values of P over time, it can be seen that the control is activated for the pitch or yaw when P exceeds 1.0 (as explained in Sec. III). The Fig~\ref{fig:area1data} demonstrates this for a particular trial in Arena 1.

\begin{figure}[h!]
  \centering
  \includegraphics[width=0.5\textwidth]{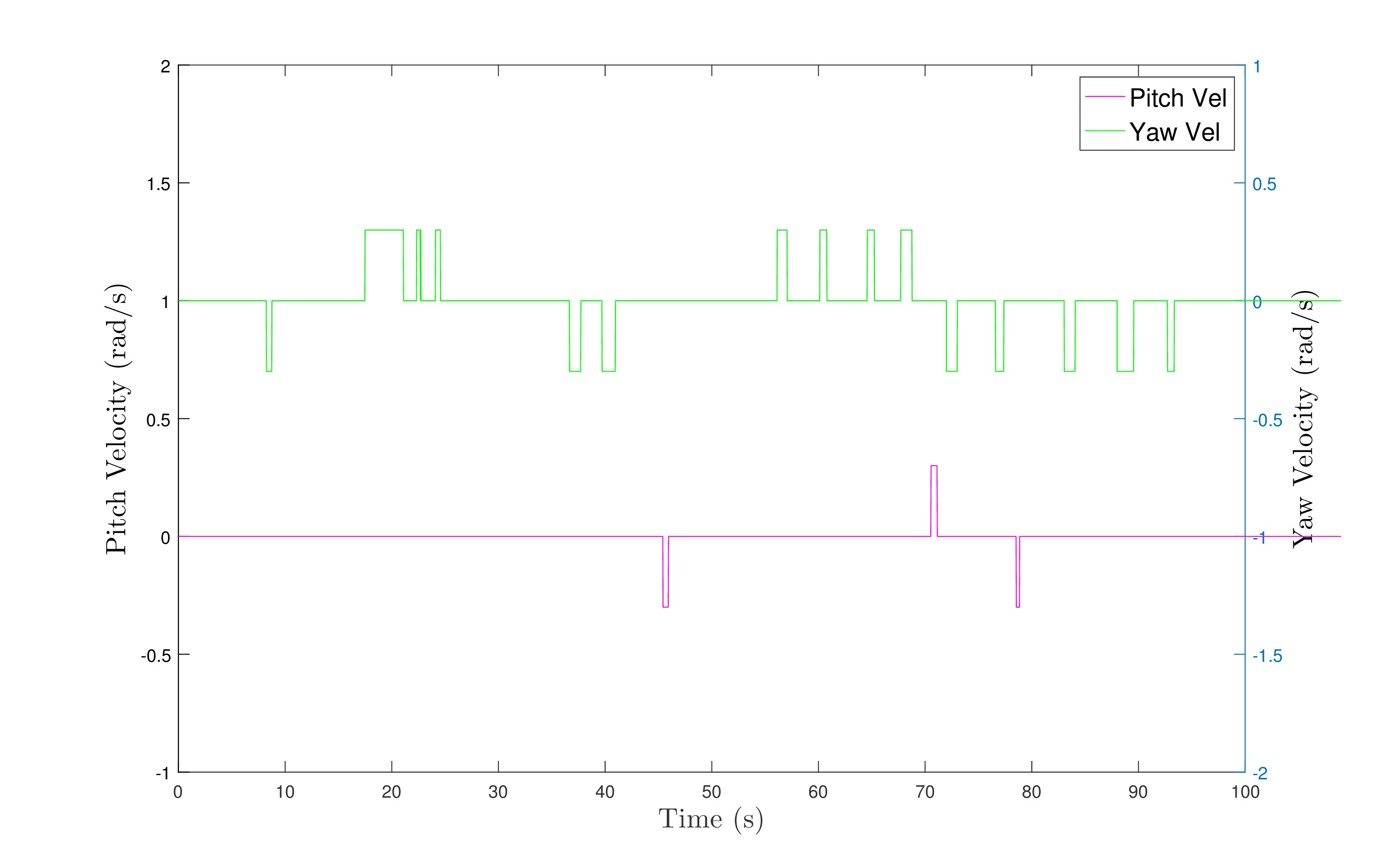}
  \caption{Pitch/Yaw control - Arena 1}
  \label{fig:area1data1}
\end{figure}

Fig~\ref{fig:area1data1} shows the control expenditure of pitch and yaw for the same test case. It can be seen that pitch and yaw motions are almost always mutually exclusive when the object moves out of the ROI, thereby reducing the number of camera movements in tracking the object.

The mean sensitivities for the 13 trials of Arena 1 is calculated to be 0.6045. 

\subsection{Arena 2}

\begin{figure}[h!]
  \centering
  \includegraphics[width=0.5\textwidth]{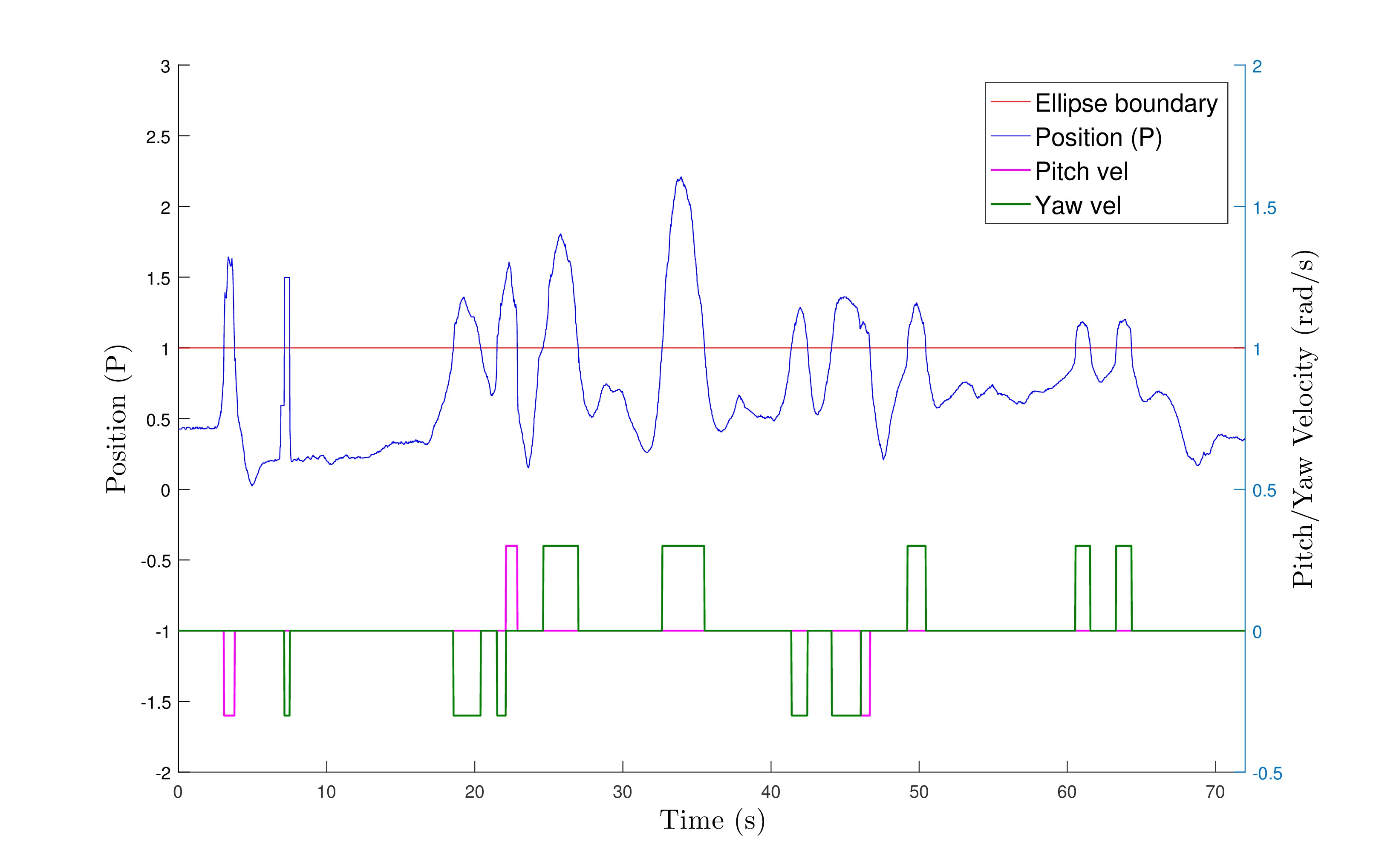}
  \caption{Time history of recorded data - Arena 2}
  \label{fig:area2data}
\end{figure}

For the second test case 17 trials were performed with 100\(\%\) success rate in tracking the red object, as in none of the trials the camera lost track of the object. Fig~\ref{fig:area2data} demonstrates this for a particular trial in Arena 2 as it can be seen that when the value of P becomes greater than 1, the controller acts and tries to bring the object inside/on the ellipse. 

The mean sensitivities for the 17 trials of Arena 2 is obtained to be 0.4536 and the normalized sensitivity value for the two arenas is 0.034. The sensitivities are low due to the relatively low value of the fixed pitch and yaw gimbal velocities of 0.3 rad/s in the Fotokite UAV, but can be significantly ramped up or stepped down by using a different setup according to the requirement.\par
\begin{figure}[h!]
  \centering
  \includegraphics[width=0.5\textwidth]{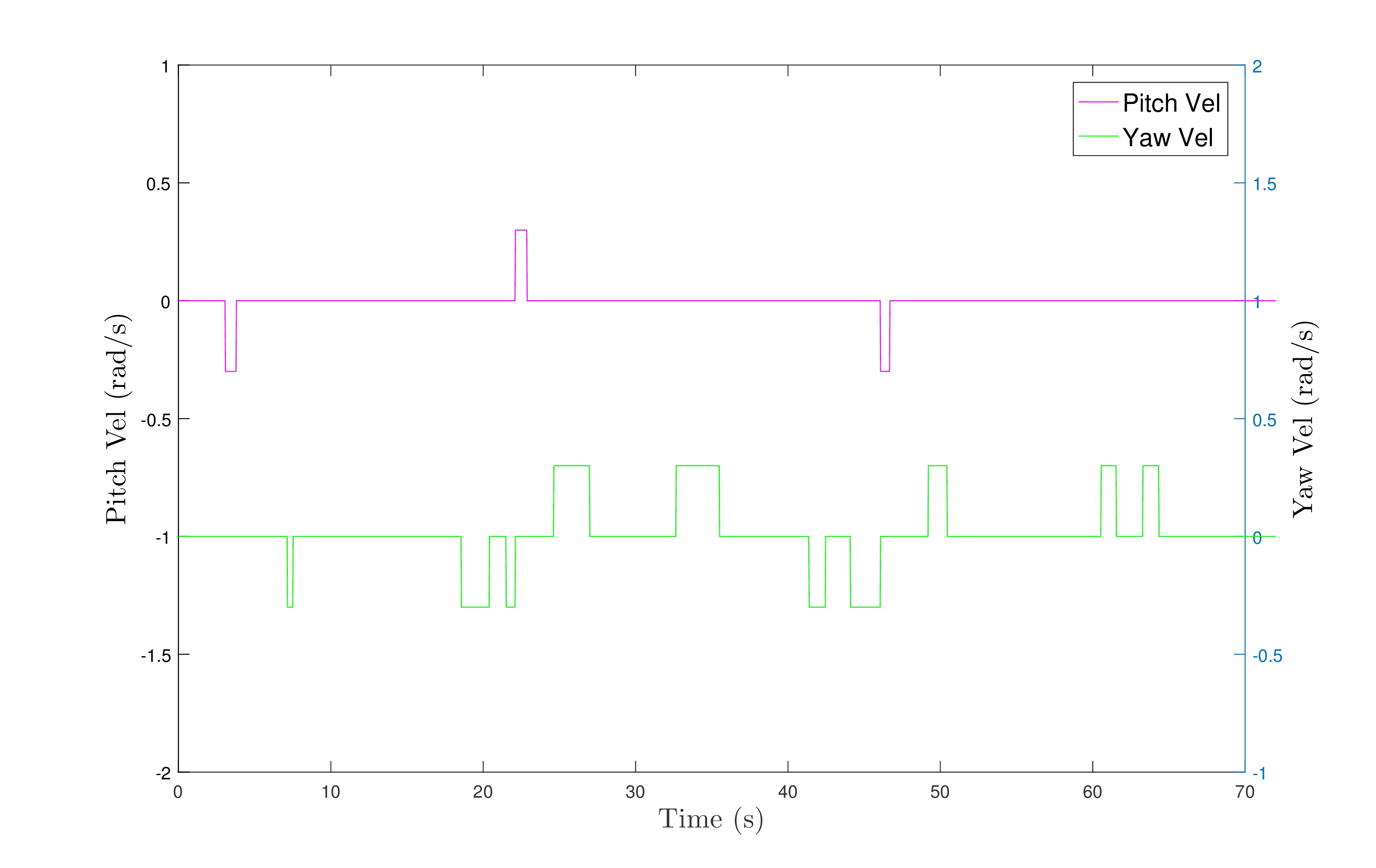}
  \caption{Pitch/Yaw control - Arena 2}
  \label{fig:area2data1}
\end{figure}

 
Fig~\ref{fig:area2data1} shows the control expenditure of the pitch and yaw for the same test case.

\section{Conclusion}

A motor schema with minimum camera movements for visual servoing of a USV from UAV was developed and tested. Metrics and graphs from repeated laboratory trials as well as computer simulation trials for the two test arenas have confirmed the viability and success rate of the algorithm in tracking the USV with minimal camera movement. In both the test arenas the success rate in tracking the object was 100\(\%\) with a normalized sensitivity value of 0.034. This work shows the possibility of eliminating the marine casuality responder's work in tracking the USV from the UAV's camera in a multi-robot USV-UAV team by efficiently automating the process.

\addtolength{\textheight}{-12cm}   % This command serves to balance the column lengths

%%%%%%%%%%%%%%%%%%%%%%%%%%%%%%%%%%%%%%%%%%%%%%%%%%%%%%%%%%%%%%%%%%%%%%%%%%%%%%%%

%%%%%%%%%%%%%%%%%%%%%%%%%%%%%%%%%%%%%%%%%%%%%%%%%%%%%%%%%%%%%%%%%%%%%%%%%%%%%%%%

%%%%%%%%%%%%%%%%%%%%%%%%%%%%%%%%%%%%%%%%%%%%%%%%%%%%%%%%%%%%%%%%%%%%%%%%%%%%%%%%
\section{Future Work}
In this research, the control algorithm was tested in the laboratory (closed environment) and computer simulation only. This needs to be taken outdoors and tested in the field with USV on water bodies and UAV on open atmosphere.

Also, an optimal control method can be formulated such that it controls both pan and tilt simultaneously to bring the USV into ROI with minimal control energy as constraints. 
\section*{ACKNOWLEDGMENT}
%%%%%%%%%%%%%%%%%%%%%%%%%%%%%%%%%%%%%%%%%%%%%%%%%%%%%%%%%%%%%%%%%%%%%%%%%%%%%%%%

We would like to thank Xuesu Xiao, Tim Woodbury, and other members of the EMILY project for providing us with the Fotokite UAV and other timely help during the testing phase of this research.

\end{document}